\documentclass{article}

\usepackage[preprint]{neurips_2025}

\usepackage[utf8]{inputenc} % allow utf-8 input
\usepackage[T1]{fontenc}    % use 8-bit T1 fonts
\usepackage{hyperref}       % hyperlinks
\usepackage{url}            % simple URL typesetting
\usepackage{booktabs}       % professional-quality tables
\usepackage{amsfonts}       % blackboard math symbols
\usepackage{nicefrac}       % compact symbols for 1/2, etc.
\usepackage{microtype}      % microtypography
\usepackage{xcolor}         % colors

\usepackage{microtype}
\usepackage{graphicx}
\usepackage{subcaption}
\usepackage{booktabs} % for professional tables

\usepackage{hyperref}
\usepackage{enumitem}
\usepackage{booktabs}
\usepackage{siunitx}
\usepackage{tabularx}
\usepackage{multirow}
\usepackage{adjustbox}
\usepackage{color}

\usepackage{algorithm}
\usepackage{algorithmic}
\usepackage{amsmath}
\usepackage{amssymb}
\usepackage{mathtools}
\usepackage{amsthm}
\usepackage{todonotes}
\usepackage{xcolor}
\usepackage{cleveref}   % For \cref references
\usepackage{amssymb}    % For mathematical symbols
\usepackage{empheq}     % For emphasized equations
\usepackage{booktabs}  % Add this to your preamble if not already there
\usepackage{graphicx}   % for \includegraphics
\usepackage{wrapfig}    % provides wrapfigure environment

\theoremstyle{plain}

\theoremstyle{definition}

\theoremstyle{remark}

\title{Reasoning Multimodal Large Language Model: Data Contamination and Dynamic Evaluation}

\author{%
  Ming Liu, Wensheng Zhang\\
  Department of Computer Science\\
  Iowa State University\\
  Ames, IA 50010 \\
  \texttt{pkulium@iastate.edu} \\
}

\begin{document}

\maketitle

\begin{abstract}
Multimodal Large Language Models (MLLMs) show impressive vision-language benchmark performance, yet growing concerns about \textit{data contamination} (test set exposure during training) risk masking true generalization. This concern extends to reasoning MLLMs, often fine-tuned via reinforcement learning from potentially contaminated base models. We propose a novel \textbf{dynamic evaluation} framework to rigorously assess MLLM generalization, moving beyond static benchmarks. Instead of perturbing inputs, we perturb the \textbf{task} itself. Using the same visual input, models are evaluated across a \textit{family of tasks} (e.g., QA, captioning, question posing, verification) to probe diverse capabilities. This \textbf{task perturbation} reveals whether model performance is robust or reliant on superficial task-specific cues. Our approach is analogous to loss landscape sharpness: models overfit or contaminated for a single task (sharp minima) falter under task shifts, unlike models with generalizable solutions (flatter minima). We developed an automated pipeline with a calibrated judge scoring open-ended generations (captions, questions) using paraphrase and corruption sampling. Applying this framework to leading image/video MLLMs on benchmarks including MME, RealWorldQA, and CVRR-ES, we analyze each model's cross-task ``ability vector''. We demonstrate that fine-tuning on simulated test data (extreme contamination) drastically sharpens task-specific performance but harms overall generalization. Our dynamic task perturbation offers deeper insights into MLLM generalization, distinguishing genuine understanding from spurious leakage or overfitting.
\end{abstract}

\section{Introduction}

\begin{figure}[ht]
    \centering
    \vspace{-1em}
    \includegraphics[
      width=0.9\textwidth,         
      height=0.5\textheight,     
      keepaspectratio            
    ]{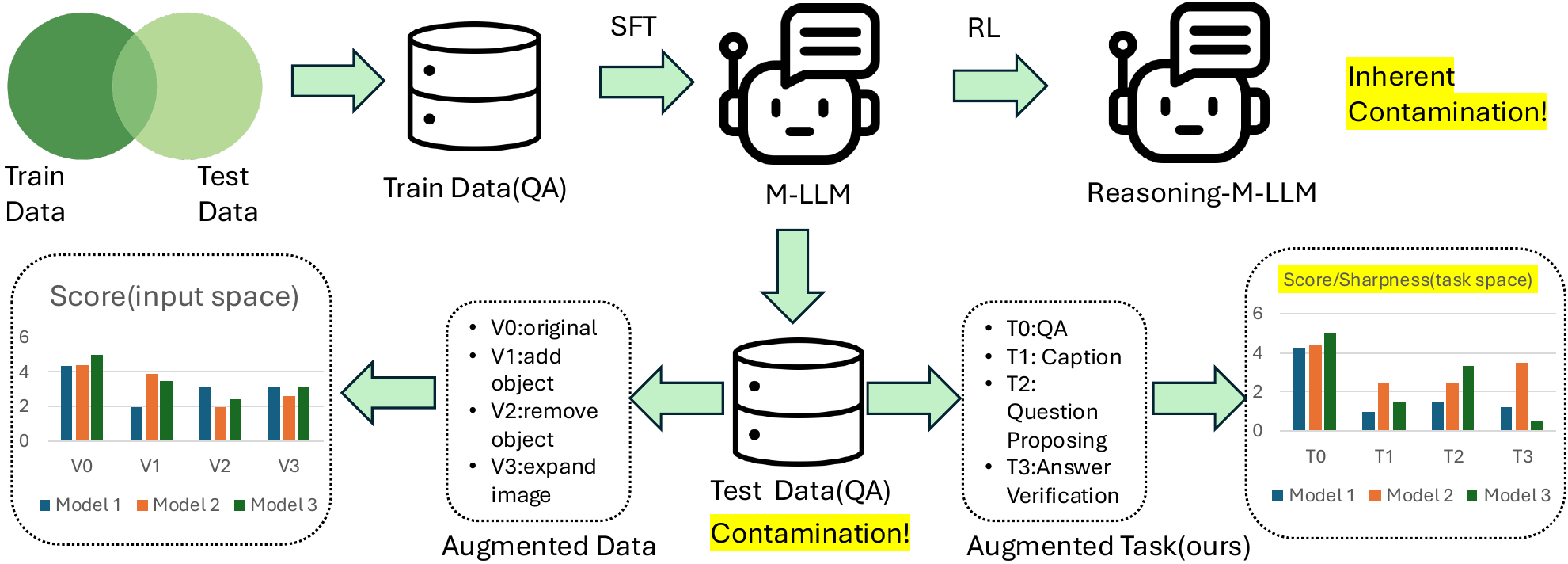}
    \caption{To mitigate contamination, we augment task and report scoring/sharpness metrics.}
    \vspace{-1em}
    \label{fig:motivation}
\end{figure}

Multimodal Large Language Models (MLLMs), combining vision and language, have advanced rapidly, with models such as GPT-4.1~\citep{openai2024gpt41} and open-source counterparts such as Qwen2-VL~\citep{qwen_team2025qwen25vl} showing strong performance in tasks such as visual question answering~\citep{fu2023mme}. More recently, reasoning-capable MLLMs, such as GPT-o4 mini~\citep{openai2024o3o4mini}, have attracted significant attention. However, the \textbf{evaluation reliability} of these MLLMs is increasingly challenged by \textit{benchmark data contamination}~\citep{chen2024we}. As shown in~\autoref{fig:motivation}, training on vast web-scale corpora makes it likely that test examples from popular benchmarks leak into training data~\citep{golchin2024time,oren2024proving}. Consequently, models may achieve high scores by \textit{memorizing} answers rather than demonstrating true understanding or reasoning, thus misrepresenting genuine generalization. This contamination extends to vision-language benchmarks~\citep{yang2025dynamic,chen2024we}. A contaminated model might correctly answer previously seen questions, undermining reported accuracy. In short, \textbf{performance gains on static benchmarks may overstate an MLLM's true generalization ability}, necessitating new evaluation methodologies.

One proposed solution is \textit{dynamic evaluation on input space}, which stress-tests models with input variations to assess true generalisation. Prior work explored input-level perturbations such as rephrasing, adding irrelevant context, or image modifications (e.g., cropping, noise) to measure consistency~\citep{zhu2024dynamic,yang2025dynamic,liu2025robustnessmultimodallanguagemodel}. However, these \textbf{input perturbation} approaches face significant limitations. \textbf{Small perturbations} (e.g., minor wording changes, slight image noise) often have negligible impact on output, particularly for robust LLM-based models adept at handling superficial prompt variations~\citep{liu2025robustnessmultimodallanguagemodel}. Memorizing contaminated models may still recognize slightly altered queries. Conversely, \textbf{large perturbations} risk altering the query's semantic intent or introducing ambiguity. For instance, substantial rewording or inserted objects in image might create a new, potentially unclear question, making performance drops difficult to interpret: is it a failure of generalization or a consequence of a changed task? It is possible that simple input modifications such as synonym replacement do not reliably preserve the original evaluation objective, and their efficacy in eliminating memorized cues lacks rigorous validation. Thus, \textbf{existing dynamic evaluation focusing on input perturbations struggles to provide a sensitive yet faithful test}: minor changes can show false stability, while major changes could make the input ambiguous.

We propose a fundamentally different paradigm for dynamic evaluation: \textbf{perturb the task, not the input}. Our insight is to hold the visual input constant, thereby preserving semantic content, while \textit{varying what we ask the model to do} with that content. Evaluating an MLLM on a \textbf{family of related tasks} for the same image or video provides a multifaceted view of its understanding. Specifically, we define four tasks probing complementary reasoning skills with the same input: \textbf{(1) Visual Question Answering (QA)}, \textbf{(2) Image/Video Captioning}, \textbf{(3) Question Generation (Question Proposing)}, and \textbf{(4) Answer Verification}. For a given visual input, the model might answer a specific question (QA), describe it freely (captioning), generate a relevant question (generation), or verify a candidate answer (verification).~\autoref{fig:tasks} conceptually illustrates this (same input, different prompts). These tasks are chosen as each requires overlapping visual knowledge but exercises distinct capabilities.

By \textbf{evaluating across this task spectrum}, we better assess MLLM generalization. A robust model with an accurate internal visual representation should perform well across \textit{all} tasks or show graceful degradation. In contrast, a model achieving high scores on a single-task benchmark (e.g., QA) via data leakage or narrow overfitting may exhibit a stark imbalance: succeeding at answering  questions but failing at coherent description or relevant question generation for the same image. In essence, \textbf{task perturbation is a stress test for model adaptability}. If a model merely \textit{matches questions to answers} by recall, this skill offers little help for \textit{generating a question} or \textit{verifying an answer}. Our dynamic evaluation thus exposes performance variations that static single-task evaluations miss.

A \textbf{theoretical perspective} explains why task perturbation reveals contamination. If a model is fine-tuned on a leaked task $t_0$ (e.g., QA), its parameters $\theta$ are near a loss minimum $L_{t_0}(\theta)$, leading to memorization and a \textbf{sharp minimum}. Small deviations from $t_0$ sharply increase error. When tested on a perturbed task $t_1$ (e.g., captioning), an overfit model shows a large performance drop and high $L_{t_1}(\theta)$, with a large gradient $\nabla_\theta L_{t_1}$. In contrast, a model with better generalisation occupies a \textbf{flatter region}, with modest increases in $L_{t_1}(\theta)$, showing stable performance. Thus, \textbf{task-space sharpness} distinguishes memorization from general capability. Our framework uses the cross-task \textbf{performance vector} $(\text{Score}_{t_0}, \text{Score}_{t_1}, \dots)$ to infer sharpness. Contaminated models show spiky profiles (high score on $t_0$, low on others), while robust models show flatter profiles. Fine-tuning on $t_0$'s test set maximizes contamination, perfecting $t_0$'s performance but degrading others, revealing the sharpness gap.

Practical implementation of this dynamic evaluation faces several challenges. First, scoring \textbf{subjective generative tasks} (captioning, question generation) is non-trivial, unlike \textbf{objective tasks} (QA, answer verification) with ground truth. We address this by using a \textbf{judge model}. An reasoning MLLM (e.g., VL-Rethinker) are used to determine response quality, enabling quantitative scoring of captions and questions proposed with reasoning process. Second, a unified per-input pipeline integrates all four task evaluations. An offline script per visual tests: (a) QA (vs. ground truth), (b) captioning (judge-scored), (c) question generation (judge-scored), and (d) answer verification (vs. varied ground-truth pairs). This yields rich per-model, per-input metrics. Finally, results are aggregated for the \textbf{cross-task analysis}.

\textbf{Contributions.} Our primary contributions are:

\begin{itemize}
\item \textbf{New Evaluation Framework:} Introduction of \textit{dynamic task perturbation}, a novel MLLM dynamic evaluation that perturbs the \textit{task} (objective), not the input, to probe generalization. This general method is extendable.
\item \textbf{Task Family Design:} Design of a four-task family (QA, captioning, question generation, answer verification) using the same visual input to test complementary reasoning and cross-task consistency, justified to reveal reliance on narrow patterns over true understanding.
\item \textbf{Automated Generative Task Evaluation:} Development of automated evaluation for subjective tasks via a \textbf{reasoning judge model} for reliable scoring.
\item \textbf{Comprehensive Evaluation:} Instantiation of the framework on image (\textbf{RealWorldQA}, \textbf{MME}) and video (\textbf{CVRR-ES}) benchmarks, with uniform evaluation of a broad suite of 12 image and 11 video state-of-the-art MLLMs (open and closed-source).
\item \textbf{Insightful Analysis (Task-Space Sharpness):} Analysis of cross-task performance (as an \textbf{ability vector}), where variance indicates specialization versus generalization (sharp vs. flat optima analogy). A controlled \textit{contamination simulation} (fine-tuning on test QA data) highlights performance profile contrasts, offering robustness insights.
\end{itemize}

Our findings validate that dynamic task perturbation is highly revealing: models known to likely have seen the benchmark during training show markedly uneven performance, whereas models with true generalisation handle task shifts better. We hope this framework will serve as a \textbf{rigorous new standard for evaluating multimodal models}, encouraging the development of models that truly \textit{reason} about visual content rather than exploiting spurious priors or leaked data.

\section{Method: Fixed-Input Multi-Task Robust Evaluation}
\label{sec:method}

Our proposed dynamic evaluation framework centers on perturbing the task rather than the input. This section details the theoretical motivation for this approach, defines the fixed-input multi-task evaluation framework, specifies the task family, introduces metrics for robustness, and outlines the practical implementation including the evaluation pipeline and the judge model for subjective tasks.

\subsection{Theoretical Motivation and Rationale}

\paragraph{Limitations of Input Perturbations}
Conventional dynamic evaluations via input perturbations (e.g., rephrased questions, image noise) present a dilemma: minor changes often fail to expose weaknesses in robust MLLMs, whereas substantial alterations can shift query semantics, rendering performance drops ambiguous. Contaminated models may recognize slightly perturbed inputs and still provide memorized answers, yielding a false sense of robustness. Our task perturbation approach circumvents this: by fixing the input and varying the task, we probe model understanding from multiple angles.

\paragraph{Task Perturbation and Optimization Landscape}
Task perturbation's efficacy in revealing model robustness and potential contamination is rooted in the optimization loss landscape concept. A model heavily trained on a specific task $t_0$, particularly with contaminated training data (leaked $t_0$ test examples), likely has parameters $\theta$ in a \textit{sharp minimum} of its loss $L_{t_0}(\theta)$. This indicates high optimization for $t_0$, but potentially in a very narrow optimum. On a different but related task $t_1$ (with the same input), such a model may show a significant performance drop, as $\theta$ is suboptimal for $L_{t_1}(\theta)$. Conversely, a model with generalized understanding (from diverse, uncontaminated data or multi-task learning) would likely occupy a \textit{flatter region} of the loss landscape, ensuring more stable performance across various tasks $t_0, t_1, \dots, t_k$. This ``task-space sharpness'' can thus indicate overspecialization or memorization versus true general capabilities.

\paragraph{Formalizing Loss Variation under Perturbation}
To formalize the sensitivity to perturbations, consider a model parameterized by $\theta$ at input $x$. If we apply a small perturbation $\delta$ in the input space, a second-order Taylor expansion of the loss $\ell(\theta; x)$ yields:
$$
\ell(\theta;\,x + \delta)
\;\approx\;
\ell(\theta;\,x)
\;+\;
\nabla_x\ell(\theta;x)^\top\,\delta
\;+\;
\tfrac12\,\delta^\top H_x(\theta)\,\delta
\;+\;O(\|\delta\|^3).
$$
Subtracting the base loss $\ell(\theta;x)$ gives the loss difference:
$$
\text{diff}(\theta; x, \delta)
=\ell(\theta;x+\delta)-\ell(\theta;x)
\approx
(\nabla_x\ell(\theta;x))^\top\delta
+\tfrac12\,\delta^\top H_x(\theta)\,\delta.
$$
The first term, $(\nabla_x\ell)^\top\delta$, captures the model’s linear sensitivity to small shifts in the input, a phenomenon extensively studied in adversarial example research (e.g., \cite{goodfellow2015explainingharnessingadversarialexamples}). The second term, $\tfrac12\,\delta^\top H_x\,\delta$, quantifies the local curvature of the loss landscape around $x$, governing how rapidly the loss can increase with moderate $\|\delta\|$.

\paragraph{Distinguishing Clean vs. Contaminated Models via Loss Landscape Differences}
When contrasting a clean (no data contamination) model with a contaminated(or over-fit) model using input perturbations, we observe differences in their loss variations:
$$
\begin{aligned}
\text{diff}_{\rm clean}(\theta; x, \delta)
&\approx\nabla_x\ell_{\rm clean}^\top\delta
+\tfrac12\,\delta^\top H_x^{\rm clean}\,\delta,\\
\text{diff}_{\rm cont}(\theta; x, \delta)
&\approx\nabla_x\ell_{\rm cont}^\top\delta
+\tfrac12\,\delta^\top H_x^{\rm cont}\,\delta.
\end{aligned}
$$
Empirically, over-fit models often satisfy $\nabla_x\ell_{\rm cont}\approx0$ for inputs they have ``memorized'', while their Hessians $H_x^{\rm cont}$ tend to exhibit higher curvature (sharper basins). The gap in their loss differences becomes:
$$
\Delta\,\text{diff}
=\text{diff}_{\rm cont}-\text{diff}_{\rm clean}
\approx
-\nabla_x\ell_{\rm clean}^\top\delta
\;+\;
\tfrac12\,\delta^\top\bigl(H_x^{\rm cont}-H_x^{\rm clean}\bigr)\,\delta.
$$
Here:
\begin{enumerate}
    \item The gradient-gap term, $-\nabla_x\ell_{\rm clean}^\top\delta$, reflects how a well-generalized model's loss might change linearly with small perturbations.
    \item The curvature-gap term, $\tfrac12\,\delta^\top(H_x^{\rm cont}-H_x^{\rm clean})\delta$, often dominates for moderate $\|\delta\|$ when $H_x^{\rm cont}$ has larger eigenvalues than $H_x^{\rm clean}$, indicating the sharper geometry of over-fit solutions.
\end{enumerate}
This decomposition suggests that moderate input perturbations $\delta$ can trigger a larger loss increase in a contaminated model compared to a clean model.

\paragraph{Connection to Flat vs. Sharp Minima Theory}
Our curvature-gap insight connects to the established flat-vs-sharp minima paradigm: \textbf{Flat minima} feature broad, low-curvature regions (in parameter or input space for a fixed model) where small changes minimally increase loss. These are often linked to better generalization (\cite{hochreiter1997flatJKU}); \textbf{Sharp minima} involve narrow basins where slight deviations rapidly escalate loss, often associated with brittle models or overfitting. Large-batch training, for instance, can converge to such minima, potentially harming generalization (\cite{keskar2016largebatch}).

Our framework extends this: rather than probing a single task with input perturbations $\delta$, we fix inputs and stress-test across different tasks. This reveals the ``flatness'' of a model's performance landscape across a spectrum of related multimodal tasks. A model excelling on one task (e.g., QA via contamination) but failing on others (e.g., captioning or question generation for the same input) thus exhibits a ``sharp'' performance profile in the task space.

\subsection{The Fixed-Input, Multi-Task Evaluation Framework}

\paragraph{Framework Definition.}
We define our evaluation framework more formally. Let:
\begin{itemize}
    \item $\mathcal{X}$ is the input space, consisting of visual inputs $X$ (images or videos). For MLLMs, the effective input for a task also includes a textual component, such as a question $Q$, so an element of $\mathcal{X}$ can be $(X, Q_{context})$, where $Q_{context}$ is task-specific.
    \item $\mathcal{Y}$ os space of possible outputs (e.g., answers, captions, generated questions, verifications).
    \item $\mathcal{T} = \{t_0, \dots, t_m\}$ be a set of $m$ distinct tasks.
    \item For each task $t \in \mathcal{T}$, a model $f_{\theta}$ (parameterized by $\theta$) effectively uses a task-specific head or prompting strategy, denoted $g_{t, \theta}$, to map an input $x \in \mathcal{X}$ (which includes the visual $X$ and any task-specific textual context) to an output $y \in \mathcal{Y}_t \subseteq \mathcal{Y}$.
    \item $\ell_t: \mathcal{Y}_t \times \mathcal{Y}_t^{gt} \to [0,1]$ be a bounded loss function for task $t$, where $\mathcal{Y}_t^{gt}$ is the space of ground-truth labels for task $t$. For subjective tasks(e.g., captioning, questions proposing), $\ell_t$ might be derived from a judge model's score.
\end{itemize}
The core principle is to evaluate $f_{\theta}$ using $g_{t,\theta}$ for all $t \in \mathcal{T}$ while keeping the primary visual input $X$ constant.

\paragraph{Proposed Task Family for Multimodal Reasoning.}
\label{sub:task_family_multimodal_reasoning}
We instantiate this framework with a four-task family probing various multimodal reasoning aspects, all grounded in the same visual input $X$:

\begin{figure}[ht]
    \centering
    \includegraphics[
      width=0.9\textwidth,          % still allow it to span the page width…
      height=0.5\textheight,     % …but never exceed half the text height
      keepaspectratio            % preserve the original aspect ratio
    ]{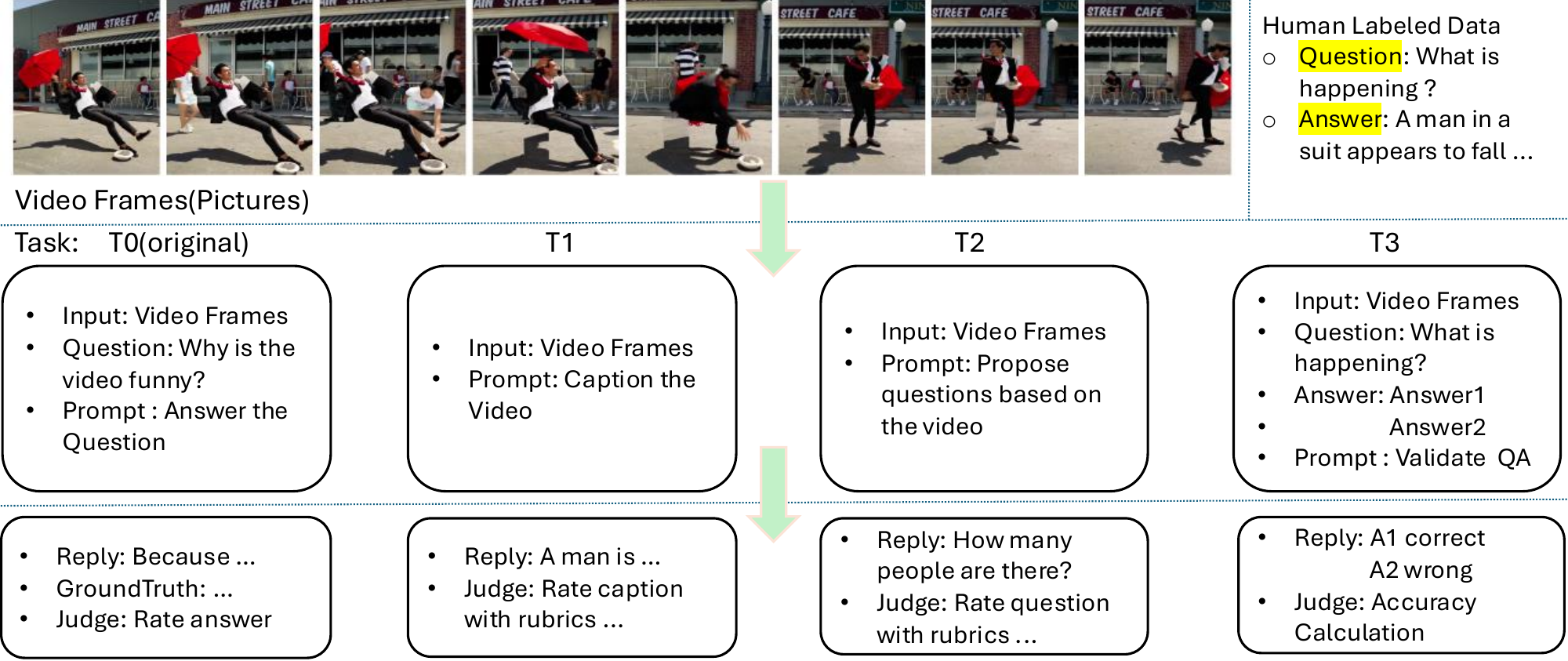}
    \caption{How to convert a QA only benchmark into multi-tasks evaluation benchmark?}
    \label{fig:tasks}
\end{figure}

\begin{itemize} 
\item \textbf{$t_0$: Visual Question Answering (QA).} Given $X$ and question $Q$ (e.g., ``Why is the video funny?''), the model provides answer $A$ (e.g., ``Because...''). This is a \textbf{discriminative} task with specific factual answers. \textbf{Evaluation:} Rate by comparing ground-truth answers. 
\item \textbf{$t_1$: Captioning.} Model generates a descriptive caption $C$ for $X$ (e.g., from a prompt like ``Caption the video.''). Example: $C$: ``A man standing...'' A \textbf{generative}, open-ended task. \textbf{Evaluation:} Caption quality scored by a judge model (Sec. \ref{sub:judge_model}). 
\item \textbf{$t_2$: Question Generation (QGen).} Prompted (e.g., ``Propose a related question $X$''), the model outputs a plausible question $Q'$ (e.g., ``How many people...''). A \textbf{generative}, underspecified task. \textbf{Evaluation:} Judge model assesses $Q'$ based on rubrics. 
\item \textbf{$t_3$: Answer Verification.} Given $X, Q,$ and candidate answer $A'$, model verifies if $A'$ is correct. E.g., $X$, $Q$: ``Is the man holding a guitar?'', $A'$: ``No'' $\rightarrow$ ``Incorrect.'' A \textbf{discriminative} task. \textbf{Evaluation:} Accuracy of these binary (correct/incorrect) judgments. 
\end{itemize}

All tasks use input $X$ but vary in query or output format. The task family thus covers factual recall ($t_0$), description ($t_1$), inquisitive reasoning ($t_2$), and critical judgment ($t_3$).

\subsection{Metrics for Assessing Robustness and Generalization}

Given a model $f_{\theta}$ and a test set of $n$ examples, we first collect its normalized performance scores into a matrix
\[
\mathbf{R}(\theta) = \bigl[R_{i,t}(\theta)\bigr] \in [0,1]^{n \times m},
\]
where
\[
R_{i,t}(\theta)
= 1 - \ell_t\bigl(g_{t,\theta}(X_i, Q_{i}^{t}), Y_i^{gt,t}\bigr),
\]
is the normalized score (e.g., accuracy or judge‐based score) of model $\theta$ on sample $i$ under task $t$, and $m=|\mathcal T|$.

\paragraph{Aggregate Risk and Worst-Task Risk}  
For each task $t$, the mean performance and corresponding risk are
\[
\widehat{P}_t(\theta)
= \frac{1}{n}\sum_{i=1}^n R_{i,t}(\theta),
\qquad
\widehat{R}_t(\theta) = 1 - \widehat{P}_t(\theta).
\]
The \emph{worst-task risk} is then
\[
R_\star(\theta)
= \max_{t}\,\widehat{R}_t(\theta)
= 1 - \min_{t}\,\widehat{P}_t(\theta),
\]
highlighting the task on which the model is weakest.

\paragraph{Inter-Task Distance Sharpness}  
To capture how differently the model behaves across tasks at the per‐sample level, let each column
\[
\mathbf{r}_t = \bigl(R_{1,t},R_{2,t},\dots,R_{n,t}\bigr)^\top
\]
be the vector of scores for task $t$. We define pairwise distances between tasks by
\[
d_{t,u}(\theta)
= \frac{1}{n}\,\bigl\lVert \mathbf{r}_t - \mathbf{r}_u \bigr\rVert_p,
\]
where $\lVert\cdot\rVert_p$ can be the Euclidean norm ($p=2$), Manhattan norm ($p=1$). Then:
\[
S_{\mathrm{dist}}(\theta)
= \max_{t,u} \; d_{t,u}(\theta),
\quad
\overline{S}_{\mathrm{dist}}(\theta)
= \frac{2}{m(m-1)} \sum_{t<u} d_{t,u}(\theta).
\]
A larger $\overline{S}_{\mathrm{dist}}$ indicates a sharper discrepancy between at least one pair of tasks.

\subsection{Practical Implementation}

\paragraph{Automated Evaluation Pipeline}
The evaluation pipeline involves iterating through test data and applying each task from the family $\mathcal{T}$ to models under evaluation.
Algorithm \ref{alg:dynamic_eval} outlines this process.

% \begin{figure}[ht]
%     \centering
%     \includegraphics[
%       width=0.7\textwidth,          % still allow it to span the page width…
%       height=0.7\textheight,     % …but never exceed half the text height
%       keepaspectratio            % preserve the original aspect ratio
%     ]{figs_tables/calibrate_judge.pdf}
%     \caption{calibrate judge}
%     \label{fig:calibrate_judge}
% \end{figure}

\paragraph{Evaluating Subjective Tasks with a Reasoning MLLM as Judge}
\label{sub:judge_model}
Evaluating free-form text from tasks $t_1$ (captioning) and $t_2$ (question generation) is challenging~\citep{liu2025is}. We use an \textbf{reasoning judge model} to assess output quality. Reasoning MLLM as judge provides not only rate but also a reasoning. In evaluation, the judge scores each generated caption/question based on a detailed rubrics. This score informs task loss $\ell_t$ (e.g., $1 - \text{judge\_score}$). The judge is fixed for all MLLM evaluations.

% Judge training data is LLM-generated (e.g., ChatGPT/GPT-4) synthetically, bootstrapped from a few human references:
% \begin{itemize}
% \item \textbf{Seed data collection:} Collect human-written captions/questions for out-of-test-set seed images/videos (e.g., from MS COCO, VATEX, or manual annotation).
% \item \textbf{Positive and negative generation:} For each seed instance with reference output $R$:
%  \begin{itemize}
% \item Generate \textbf{Paraphrases} of $R$ preserving meaning (positive examples).
%  \item Generate \textbf{Corrupted outputs} with typical errors (e.g., incorrect objects, irrelevant/trivial questions) (negative examples).
%  \end{itemize}
% \item \textbf{Judge model training:} Fine-tune a pretrained multimodal model on this synthetic data. Judge input: visual $X$ and candidate text (caption/question). Output: P("Positive").
% \end{itemize}

\section{Experiment}
\label{sub:contamination_amplification}
 
\subsection{Datasets and Models}
We apply our framework to three benchmarks, encompassing over 20 models. For the image benchmarks, we include RealWorldQA~\cite{realworldqa2024} and MME~\cite{fu2023mme}; for the video benchmark, we include CVRR-ES~\cite{evalai2024cvrres}. We evaluate a diverse set of MLLMs. Image-focused models include InternVL v2~\cite{chen2024internvl2}, LLaVA-HF~\cite{liu2023visual}, LLaVA-OneVision~\cite{li2025llavaonevision}, Phi-4 Multimodal~\cite{abouelenin2025phi4}, Qwen2-VL (7B)~\cite{wang2024qwen2vl}, Qwen2.5-VL (3B)~\cite{qwen_team2025qwen25vl}, Llama-Vision~\cite{llama_team2024llama3herd}, GPT-4.1~\cite{openai2024gpt41}, GPT-o4 mini~\cite{openai2024o3o4mini}, VL-Rethinker~\cite{wang2025vlrethinker}, and MM-Eureka~\cite{meng2025mmeureka}. The last three models are reasoning model. Corresponding video-capable extensions and models comprise InternVideo v2.5~\cite{wang2025internvideo2.5}, InternVL v2~\cite{chen2024internvl2}, LLaVA-OneVision~\cite{li2025llavaonevision}, Phi-4 Multimodal~\cite{abouelenin2025phi4}, Qwen2-VL~\cite{wang2024qwen2vl}, Qwen2.5-VL~\cite{qwen_team2025qwen25vl}, GPT-4.1~\cite{openai2024gpt41}, GPT-o4 mini~\cite{openai2024o3o4mini}, Video-Chat R1~\cite{li2025videochatr1}, and Video R1~\cite{feng2025videor1}. The last three models are reasoning model as well. All experiments were conducted on machine equipped with NVIDIA A100-80GB GPUs. Depending on the length of the models’ responses, each evaluation may take several hours.

\subsection{Experimental Results and Analysis}
\label{sec:results_and_analysis} % New label for the merged section

To empirically validate our task perturbation framework's ability to assess model generalization and to illustrate the concept of task-space sharpness, we first evaluate a diverse set of pretrained MLLMs on our defined family of four tasks using the MME benchmark~\citep{fu2023mme}. The detailed performance metrics for each model across tasks are presented in Table~\ref{tab:mme}. Results for RealWorldQA and CVRR-ES datasets are in~\autoref{app:more_results}. The results from Table~\ref{tab:mme} allow for an analysis of task-space sharpness and model-specific ability vectors:

\begin{table}[ht]
\centering
\caption{Performance of pretrained (reasoning) MLLMs Across Four Tasks on MME dataset.}
\label{tab:mme}
\renewcommand{\arraystretch}{1.1}
\setlength{\tabcolsep}{5pt}
\begin{tabular}{l r r r r | r r r r}
\toprule
\textbf{Model} & \textbf{T0} & \textbf{T1} & \textbf{T2} & \textbf{T3} & \textbf{Avg} & \textbf{W.Risk} & \textbf{SD} & \textbf{Rng} \\
\midrule
InternVL v2       & 77.46 & 83.91 & 77.33 & 55.31 & 73.50 & 44.69 & 10.84 & 28.60 \\
Llama-Vision      & 72.37 & 76.59 & 77.55 & 62.55 & 72.27 & 37.45 &  5.94 & 14.99 \\
LLaVA-HF          & 78.31 & 68.03 & 65.84 & 63.31 & 68.87 & 36.69 &  5.70 & 15.00 \\
LLaVA-OneVision   & 84.67 & 86.02 & 86.30 & 68.56 & 81.38 & 31.44 &  7.43 & 17.75 \\
Phi-4 MM          & 77.80 & 74.88 & 82.32 & 59.63 & 73.66 & 40.37 &  8.52 & 22.70 \\
Qwen2-VL          & 88.25 & 88.11 & 80.14 & 59.58 & 79.02 & 40.42 & 11.69 & 28.66 \\
Qwen2.5-VL        & 86.86 & 90.93 & 85.28 & 75.57 & 84.66 & 24.43 &  5.64 & 15.36 \\
GPT-4.1           & 77.42 & 99.06 & 95.77 & 82.18 & 88.61 & 22.58 &  9.04 & 21.64 \\
\midrule[\lightrulewidth]
MM-Eureka         & 90.27 & 84.84 & 73.32 & 96.04 & 86.12 & 26.68 &  8.39 & 22.73 \\
VL-Rethinker      & 87.36 & 86.19 & 83.55 & 97.54 & 88.66 & 16.45 &  5.31 & 13.99 \\
GPT-o4 mini       & 88.29 & 99.44 & 92.38 & 82.47 & 90.65 & 17.53 &  6.18 & 16.97 \\
\bottomrule
\end{tabular}
\vspace{0.5ex}
\begin{flushleft}
\footnotesize{\textit{Note:} Values shown as percentages ({\%} symbol omitted). T0–T3 = Tasks 0–3; Avg = Average Performance; W.Risk = Worst Risk; SD =  $S_{\mathrm{dist}}$; Rng = Range. Last three models are reasoing model.}
\end{flushleft}
\end{table}

\begin{figure}[ht]
    \centering
    \vspace{-1em}
    \includegraphics[width=0.7\textwidth]{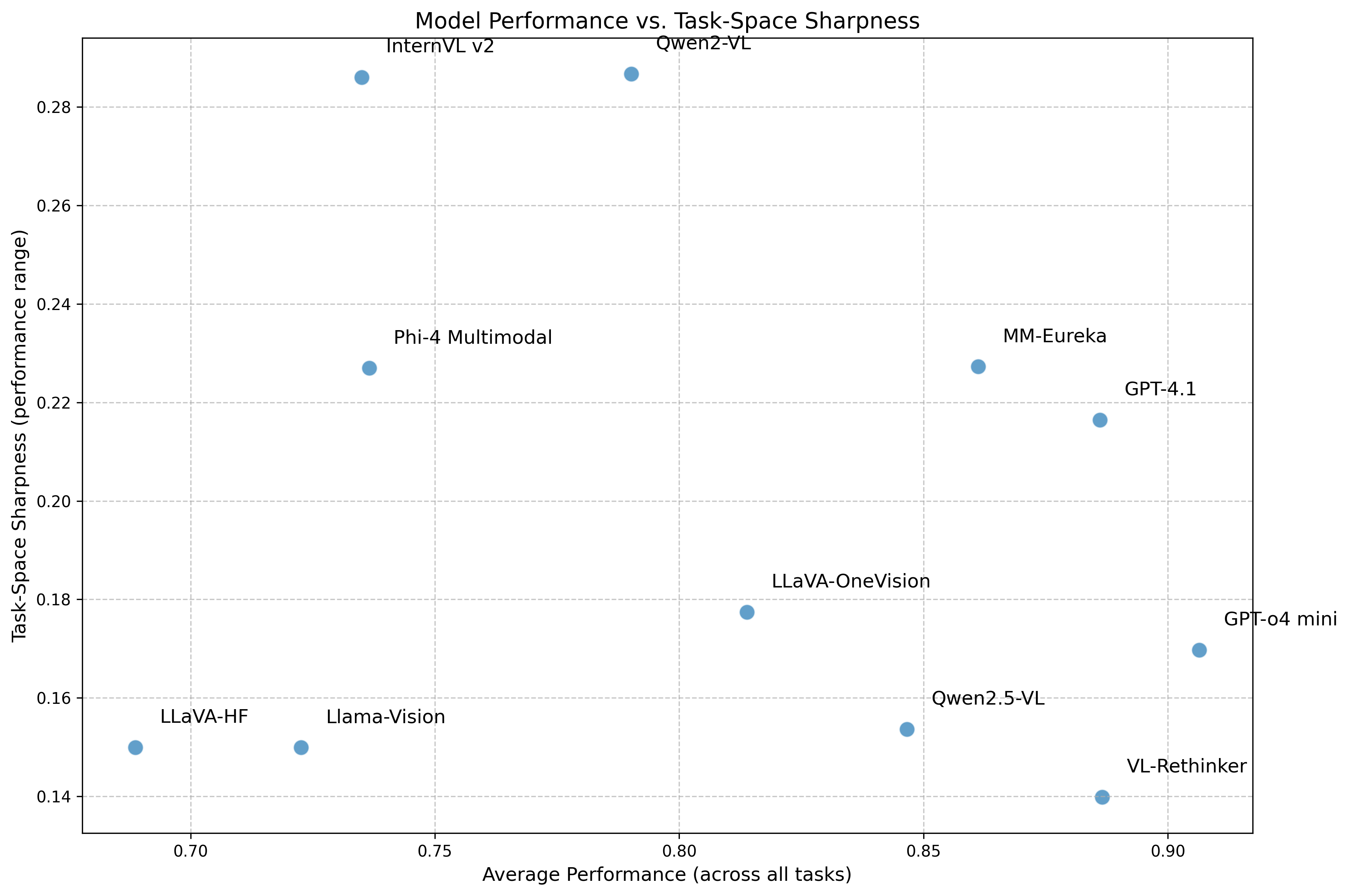}
    \caption{Model average performance versus task-space sharpness (performance range) on the MME dataset. Models in the bottom-right quadrant exhibit both high average performance and low sharpness, indicating better generalization.}
    \label{fig:sharpness_vs_performance_mme}
    \vspace{-2em}
\end{figure}

\begin{wrapfigure}[19]{r}{0.45\textwidth}
    \centering
    \vspace{-1em}
    \includegraphics[width=0.5\textwidth]{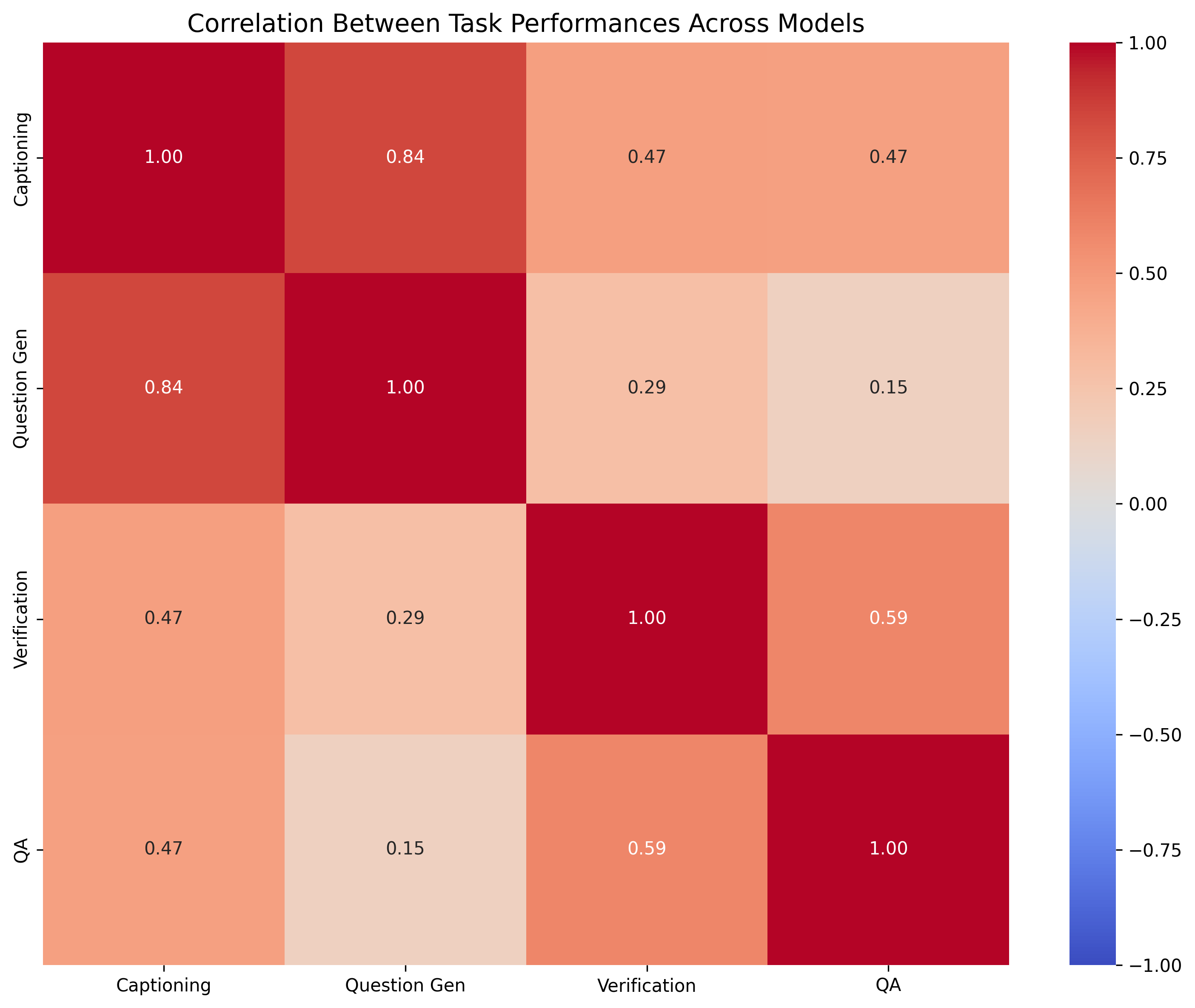}
    \caption{Correlation matrix of task performances (T0: QA, T1: Captioning, T2: Question Gen, T3: Verification) across models on the MME dataset.}
    \label{fig:task_correlation_mme}
\end{wrapfigure}

\textbf{1. Measuring Task-Space Sharpness:} Based on each model's performance vector $[P_{QA}, P_{Cap}, P_{QGen}, P_{Verif}]$, we primarily measure task-space sharpness using the \textbf{performance range (Rng)} (Table~\ref{tab:mme}). Rng directly quantifies $S_{dist}(\theta)$ (maximum pairwise task distance), capturing the maximum performance discrepancy and is visualized in our figures. We also report the \textbf{standard deviation (SD)} of task scores, corresponding to $\overline{S}_{dist}(\theta)$ (average pairwise task distance). A large Rng (e.g., InternVL~v2) signifies high sharpness, possibly from contamination or narrow training, while a smaller Rng (e.g., GPT-o4~mini) indicates balanced performance and stronger generalization.

Figure~\ref{fig:sharpness_vs_performance_mme} plots the average cross-task performance against task-space sharpness (performance range) for the evaluated models on the MME dataset. As hypothesized, models demonstrating stronger generalization, such as GPT-o4 mini (Avg: $90.65\%$, Rng: $16.97\%$) and VL-Rethinker (Avg: $88.66\%$, Rng: $13.99\%$), tend to occupy the desirable \textbf{bottom-right region} of the plot. This positioning indicates both high overall capability and low performance variance across tasks. Conversely, models like InternVL v2 (Avg: $73.50\%$, Rng: $28.60\%$) and Qwen2-VL (Avg: $79.02\%$, Rng: $28.66\%$) exhibit higher sharpness values for their respective performance levels, suggesting a greater imbalance in their capabilities across the evaluated tasks.

Further analysis of Table~\ref{tab:mme} reveals notable \textbf{QA-to-Caption gaps} and \textbf{QA-to-Verification gaps} for certain models. For instance, InternVL v2 scores $77.46\%$ on T0 (QA) but only $55.31\%$ on T3 (Verification), a significant drop. Similarly, Qwen2-VL achieves $88.25\%$ on T0 but $59.58\%$ on T3. These discrepancies suggest that high QA performance might not always translate to comprehensive visual understanding or reliable verification, potentially stemming from memorization or overfitting to QA-specific cues rather than holistic reasoning.

\textbf{2. Interpreting Model Ability Vectors:} By comparing the ability vectors of different models (rows in Table~\ref{tab:mme}), we can identify patterns:
\begin{itemize}
    \item \textbf{Reasoning Model:} Models with RL training strategies may exhibit comparable task profiles. For example, reasoning-focused models(RL trained) like VL-Rethinker and GPT-o4 mini generally show strong, balanced performance. This aligns with prior findings~\citep{chu2025sftmemorizesrlgeneralizes}, showing that SFT tends to memorize while RL promotes generalization.
    \item \textbf{Uniformly Strong Generalists:} Models like GPT-o4 mini demonstrate consistently high scores across all four tasks, positioning them as strong generalists on this benchmark. 
    GPT-4.1 also shows high average performance.
    \item \textbf{Task-Specific Strengths/Weaknesses:} Some models exhibit clear preferences. For instance, Qwen2-VL excels at T0 (QA, $88.25\%$) and T1 (Captioning, $88.11\%$) but shows considerably lower performance on T3 (Verification, $59.58\%$). LLaVA-HF, conversely, has a relatively lower T1 score ($68.03\%$) compared to its T0 score ($78.31\%$). These variations can offer insights into the specific training data or objectives these models were optimized for.
    \item \textbf{Task Correlations:} Figure~\ref{fig:task_correlation_mme} presents the correlation matrix of task performances on the MME dataset. Notably, Captioning (T1) and Question Generation (T2) exhibit a strong positive correlation ($r=0.84$), suggesting that models proficient in captioning an image are often also adept at formulating relevant questions for it. Captioning (T1) also shows a moderate positive correlation with QA ($r=0.47$). The correlation between Question Verification (T2) and Question Generation (T3) is weaker ($r=0.29$) on this dataset for the evaluated models, which might imply that verifying a given answer leverages a somewhat different skill set than proposing questions, or it could highlight model inconsistencies in these areas.
\end{itemize}

Further, to directly demonstrate how our framework can expose overfitting and the effects of data contamination, we conduct a controlled experiment. We select a subset of models (Qwen2.5-VL 3B and 7B) and fine-tune them using Parameter-Efficient Fine-Tuning (PEFT)~\citep{hu2021loralowrankadaptationlarge} directly on the test Question Answering (T0) data from the RealWorldQA benchmark. This procedure simulates an extreme case of data contamination specifically for the QA task.

\begin{table}[htbp]
\centering
\caption{Performance of pretrained and extreme contaminated models(PEFT finetuned on test data with QA task T0) across four tasks.}
\label{tab:compare_clean_with_contaminated_model}
\renewcommand{\arraystretch}{1.1}
\setlength{\tabcolsep}{5pt}
\begin{tabular}{l r r r r | r r r r}
\toprule
\textbf{Model} & \textbf{T0} & \textbf{T1} & \textbf{T2} & \textbf{T3} & \textbf{Avg} & \textbf{W.Risk} & \textbf{SD} & \textbf{Rng} \\
\midrule
Qwen2.5-VL (3B)      & 58.82 & 88.50 & 70.78 & 80.42 & 74.63 & 41.18 & 11.07 & 29.68 \\
Qwen2.5-VL (7B)      & 63.27 & 91.58 & 64.64 & 89.43 & 77.23 & 36.73 & 13.31 & 28.31 \\
\midrule[\lightrulewidth]
Qwen2.5-VL+PEFT (3B) & 96.21 & 85.16 & 63.40 & 81.43 & 81.55 & 36.60 & 11.80 & 32.81 \\
Qwen2.5-VL+PEFT (7B) & 96.21 & 90.75 & 69.08 & 89.18 & 86.31 & 30.92 & 10.28 & 27.12 \\
\bottomrule
\end{tabular}
\vspace{0.5ex}
\begin{flushleft}
\footnotesize{\textit{Note:} PEFT indicates models fine-tuned on test dataset using Parameter-Efficient Fine-Tuning.}
\end{flushleft}
\end{table}

As shown in Table~\ref{tab:compare_clean_with_contaminated_model}, this fine-tuning indeed leads to a dramatic improvement in T0 (QA) performance for the contaminated models. Crucially, we also observe corresponding shifts in performance across other tasks (T1--T3) and notable changes in task-space sharpness metrics (Range and SD). This setup vividly demonstrates how task perturbation can expose overfitting that a static, single-task evaluation might miss, as the contaminated model would appear state-of-the-art on T0.

The impact of this simulated data contamination (fine-tuning on test data) is starkly illustrated by comparing the original Qwen2.5-VL models with their PEFT-tuned counterparts as shown in Table~\ref{tab:compare_clean_with_contaminated_model}:
\begin{itemize}
    \item For Qwen2.5-VL (3B), T0 (QA) performance dramatically increases from $58.82\%$ to $96.21\%$ after PEFT fine-tuning. However, its performance on T1 (Captioning) slightly decreases from $88.50\%$ to $85.16\%$, and T2 (Question Generation) drops from $70.78\%$ to $63.40\%$. This specialization leads to an increase in task-space sharpness, with the performance range (Rng) widening from $29.68\%$ to $32.81\%$, and the standard deviation (SD) also increasing from $11.07\%$ to $11.80\%$.
    \item For the larger Qwen2.5-VL (7B) model, T0 performance similarly spikes from $63.27\%$ to $96.21\%$. Its performance on T1 (Captioning) sees a minor decrease ($91.58\%$ to $90.75\%$), and T3 (Verification) also slightly drops ($89.43\%$ to $89.18\%$). The critical observation is the \textit{induced imbalance}: the model becomes almost exclusively proficient at the contaminated task, with other capabilities not showing commensurate improvement or even degrading.  
    \item This dramatic change illustrates how optimizing for a single task via test set exposure can adversely affect, or at least not proportionally benefit, performance on related tasks, reaffirming the need for multi-task evaluation to get a holistic view of model capabilities.
\end{itemize}

% \textbf{Case Studies:} Lastly, while quantitative metrics provide a broad overview, qualitative case studies are invaluable for understanding \textit{how} models succeed or fail when tasks are perturbed. We include examples from RealWorldQA and CVRR-ES illustrating instances where a model correctly answer a QA query (T0), perhaps by relying on spurious correlations, but then provide a nonsensical caption (T1) or generate an irrelevant question (T2) for the same visual input. Such examples provide concrete evidence of the diagnostic power of task perturbation, offering insights into \textit{what} the models struggle with when the task demands shift.

Through this analysis, we aim to paint a clear picture of each model's generalisation. Rather than relying on a single accuracy number, our framework provides a \textbf{profile} of capabilities. This is especially valuable, since it can inform where a model might need improvement (e.g., a model proficient at description but poor at verification might require training on consistency checking), and it can reveal whether high benchmark scores are trustworthy or indicative of narrow specialization.

\section{Related Work}
\label{sec:related_work}

\textbf{Data Contamination.}
The integrity of evaluation benchmarks for both LLMs and MLLMs is increasingly challenged by data contamination~\citep{golchin2024time,chen2024we}. Significant overlap between benchmark datasets and common pretraining corpora like C4 has been documented, with exact match rates observed between 2\% and 50\% across different benchmarks~\citep{dodge2021documenting}. For instance, the Llama-2 model development revealed that over 16\% of MMLU benchmark examples were present in training data, with 11\% considered severely contaminated~\citep{touvron2023llama}. This concern is not limited to LLMs; it also affects MLLMs. Research by~\cite{chen2024we} pointed out potential contamination in multimodal benchmarks, noting that Sphinx-X-MoE~\citep{gao2024sphinx} achieved a 43.6\% score on MMMU~\citep{yue2024mmmu} using only text inputs, significantly outperforming its base LLM (17.9\%). Previous study~\citep{yang2025dynamic} reveal substantial contamination levels in both visual and textual components when comparing static multimodal benchmarks against dynamical training datasets. This widespread contamination underscores the limitations of static benchmarks, particularly when evaluating performance based on a single task format where memorization can inflate scores, and motivates the exploration of dynamic evaluation methodologies for more reliable MLLM assessment.

\textbf{MLLM Evaluation.}
The rapid advancement of MLLMs has prompted the creation of numerous benchmarks designed to assess their capabilities. Initial benchmarks often focused on specific, coarse-grained tasks, such as those represented by ok-VQA~\citep{marino2019ok} and MS-COCO~\citep{chen2015microsoft}. More recently, efforts have shifted towards developing benchmarks that offer a more comprehensive evaluation of MLLMs' overall reasoning abilities. Examples of such holistic benchmarks include MME~\citep{yin2023survey}, SEEDBench~\citep{li2023seed}, MM-Vet~\citep{yu2023mm}, MMBench~\citep{liu2023mmbench}, and Llavebench~\citep{liu2024visual}, RealWorldQA~\citep{realworldqa2024}, CVRR-ES~\citep{evalai2024cvrres}. Despite their utility, a primary drawback of these benchmarks is their static nature. Models may achieve high scores via contamination or overfitting to the specific task format presented, without necessarily demonstrating robust, generalizable understanding~\citep{chen2024we,yang2025dynamic}. This susceptibility confines evaluation to a predetermined complexity and task objective, highlighting the need for approaches that test generalization more rigorously across different facets of reasoning.

\textbf{Dynamic Evaluation.}
Dynamic evaluation is an emerging research area aimed at overcoming the limitations of static benchmarks. Several pioneering studies have explored dynamic evaluation primarily within the NLP domain. These approaches often focus on generating novel test \textit{instances} or applying \textit{input perturbations}. For example, DyVal~\citep{zhu2023dyval} was proposed to generate new test instances dynamically using graph structures, specifically to combat data contamination. NPHardEval~\citep{fan2023nphardeval} takes a similar approach by creating novel evaluation samples for NP-hard mathematical problems. Recently, studies have augmented both textual and visual inputs to dynamically evaluate MLLMs~\citep{liu2025robustnessmultimodallanguagemodel,yang2025dynamic}, although they focus solely on image-based MLLMs. While valuable, methods relying on input perturbations face the dilemma that minor changes might not affect robust models, while significant changes may risk altering the core semantics of the evaluation query. Furthermore, data augmentation can be time-consuming and resource-intensive, especially for video datasets.

Our work introduces a distinct paradigm for dynamic evaluation in the multimodal context: \textbf{task perturbation}. Instead of altering the input data or generating new instances, we maintain the original visual input (image or video) while systematically varying the \textit{task} the model is asked to perform (e.g., answer a question, generate a caption, propose a question, verify an answer). This strategy directly probes the model's adaptability and the robustness of its internal representation, assessing whether its capabilities generalize across different reasoning demands for the same core visual information. To our knowledge, this is the first work to systematically leverage task perturbation as a dynamic evaluation strategy for MLLMs, offering a complementary perspective to input-centric dynamic methods by focusing on the flexibility and consistency of model reasoning across diverse objectives.

\section{Conclusion and Limitations}

We introduce a dynamic evaluation framework for multimodal LLMs based on \emph{task perturbation}, probing generalization across QA, captioning, question generation and answer verification on the same inputs. By formalizing metrics such as worst-task risk and inter-task sharpness and grounding our analysis in loss-landscape sharp vs.\ sharp minima theory, we distinguish genuine reasoning from prompt-specific overfitting. Experiments on RealWorldQA, MME and CVRR-ES reveal hidden weaknesses and contamination risks that static benchmarks overlook, highlighting the importance of consistent performance across related tasks. This framework can be extended to new task families (e.g.\ explanation generation, dialogue), enriched with human-in-the-loop judgments, and applied to other modalities (audio, 3D). By shifting the focus from raw benchmark scores to \emph{how} and \emph{where} models succeed or fail, we aim to foster more trustworthy MLLMs. Further limitations are detailed in \autoref{app:limitations}.

\bibliography{references}
\bibliographystyle{plain}

%%%%%%%%%%%%%%%%%%%%%%%%%%%%%%%%%%%%%%%%%%%%%%%%%%%%%%%%%%%%

\appendix

\newpage
\section{Limitations}
\label{app:limitations}

While our dynamic task perturbation framework offers valuable insights into MLLM generalization, we acknowledge several limitations that suggest avenues for future research:

\begin{itemize}
    \item \textbf{Combined Perturbations:}
     Our framework currently evaluates input and task perturbations separately. Future work could combine these dimensions, for instance, by tasking a model to caption an image with deliberately introduced distractor objects.  Such composite perturbations could offer deeper insights into model robustness and fine-grained understanding under multifaceted distribution shifts.

    \item \textbf{Expanded Task Repertoire:}
     The current four-task family (QA, captioning, question generation, verification), while diverse, is not exhaustive.  Incorporating a broader range of task transformations, such as visual cloze-style (fill-in-the-blank) tasks or comparative reasoning (e.g., articulating differences between original and augmented images), could reveal different MLLM capabilities and allow for a more comprehensive assessment of reasoning skills.

    \item \textbf{Reasoning MLLMs and Inherent Contamination:}
    Evaluating "reasoning MLLMs," particularly those fine-tuned via Reinforcement Learning (RL), presents a nuanced data contamination challenge.  Since these models often derive from base MLLMs potentially exposed to benchmark data, a form of "inherent contamination" might persist despite RL fine-tuning.  Although RL training can significantly modify internal model knowledge  , the extent to which it fully mitigates prior contamination, especially subtle effects, warrants more rigorous, targeted analysis, including methods to trace pre-RL data influence on post-RL multi-task performance.

    \item \textbf{MLLM-based Judging for Subjective Tasks:}
     Employing a reasoning MLLM as a proxy judge for subjective tasks like question proposing, while scalable, carries potential empirical risks.  The MLLM judge's evaluations may reflect inherent biases, misunderstand nuanced quality aspects, or show training data artifacts, leading to discrepancies with human assessment.  Furthermore, our current protocol for question proposing evaluates the generated question's quality but does not require the model under test to subsequently answer its own proposed question. A self-consistency check could offer a more robust measure of understanding.

    \item \textbf{Manual Task Augmentation:}
    The process of defining task variants is currently manual. While this allows precise control, it is labor-intensive and may limit the diversity and scale of explored task perturbations. Future iterations could use automated agents or generative models to create a wider, more varied set of task augmentations, enhancing the framework's scalability and its ability to uncover a broader spectrum of model capabilities and failure modes.

\end{itemize}

\newpage
\section{Algorithm}
\label{app:algorithm}

\begin{algorithm}
\caption{Fixed-Input Multi-Task Evaluation Pipeline}
\label{alg:dynamic_eval}
\begin{algorithmic}[1]
\REQUIRE Model $f_{\theta}$, Task set $\mathcal{T} = \{t_0, \dots, t_m\}$, Test set $D_{test} = \{(X_i, \{Q_{context,i}^t\}_{t\in\mathcal{T}}, \{Y^{gt,t}_i\}_{t\in\mathcal{T}})\}_{i=1}^n$
\STATE Initialize $\text{losses}[t][i]$ for $t \in \mathcal{T}, i \in \{1, \dots, n\}$
\STATE Initialize $\widehat{R}_t$ for $t \in \mathcal{T}$
\FOR{each task $t \in \mathcal{T}$}
    \FOR{$i = 1 \text{ to } n$}
        \STATE Obtain visual input $X_i$ and task-specific context $Q_{context,i}^t$ (if any) from $D_{test}$
        \STATE Generate model output $\hat{y}_i^t \leftarrow g_{t,\theta}(X_i, Q_{context,i}^t)$
        \STATE Obtain ground truth $Y^{gt,t}_i$ from $D_{test}$\COMMENT{This is for tasks with groundtruth}
        \STATE Calculate loss: $\text{losses}[t][i] \leftarrow \ell_t(\hat{y}_i^t, Y^{gt,t}_i)$
            \COMMENT{For task without groundtruth, $\ell_t$ may involve the judge model}
    \ENDFOR
    \STATE $\widehat{R}_t \leftarrow \frac{1}{n}\sum_{i=1}^n \text{losses}[t][i]$
\ENDFOR
\STATE $\widehat{R}_\star \leftarrow \max_{t \in \mathcal{T}} \widehat{R}_t$ \COMMENT{Worst-task risk}
\STATE For each pair $(t,u)$ compute 
  $d_{t,u} \leftarrow \tfrac{1}{n}\,\lVert \mathbf{r}_t - \mathbf{r}_u\rVert_p$
\STATE $S_{\mathrm{dist}} \leftarrow \max_{t,u}\,d_{t,u}$ 
  \COMMENT{Maximum inter-task distance}
\STATE $\overline{S}_{\mathrm{dist}} 
  \leftarrow \tfrac{2}{m(m-1)}\sum_{t<u}d_{t,u}$
  \COMMENT{Average inter-task distance}
\STATE \textbf{return} 
  $\{\widehat{R}_t\}_{t\in\mathcal{T}},\;\widehat{R}_\star,\;S_{\mathrm{dist}},\;\overline{S}_{\mathrm{dist}}$
\end{algorithmic}
\end{algorithm}

\newpage
\section{More Results}
\label{app:more_results}

\subsection{Results for RealWorldQA dataset}
\begin{table}[h]
\centering
\caption{Performance of MLLMs Across Four Tasks on RealworldQA dataset.}
\label{tab:realworldqa}
\renewcommand{\arraystretch}{1.0}
\begin{tabular}{l r r r r | r r r r}
\toprule
\textbf{Model} & \textbf{T0} & \textbf{T1} & \textbf{T2} & \textbf{T3} & \textbf{Avg} & \textbf{W.Risk} & \textbf{SD} & \textbf{Rng} \\
\midrule
InternVL v2     & 41.70 & 75.17 & 80.31 & 68.04 & 66.31 & 58.30 & 14.86 & 38.61 \\
Llama-Vision    & 56.60 & 70.73 & 83.83 & 69.61 & 70.19 & 43.40 &  9.63 & 27.22 \\
LLaVA-HF        & 56.47 & 65.85 & 78.07 & 57.06 & 64.36 & 43.53 &  8.74 & 21.60 \\
LLaVA-OneVision & 66.67 & 85.85 & 92.18 & 54.64 & 74.83 & 45.36 & 14.97 & 37.54 \\
Phi-4 MM        & 60.13 & 76.94 & 84.11 & 65.69 & 71.72 & 39.87 &  9.38 & 23.98 \\
Qwen2.5-VL (3B) & 64.05 & 89.79 & 91.71 & 64.44 & 77.50 & 35.95 & 13.27 & 27.66 \\
Qwen2-VL (7B)   & 61.83 & 84.70 & 88.61 & 65.56 & 75.17 & 38.17 & 11.64 & 26.78 \\
GPT-4.1         & 71.11 & 90.99 & 98.39 & 79.67 & 85.04 & 28.89 & 10.44 & 27.28 \\
\midrule[\lightrulewidth]
MM-Eureka       & 45.23 & 90.81 & 81.91 & 84.51 & 75.62 & 54.77 & 17.84 & 45.58 \\
VL-Rethinker    & 67.58 & 94.84 & 90.94 & 75.75 & 82.28 & 32.42 & 11.08 & 27.25 \\
GPT-o4 mini     & 77.12 & 88.68 & 94.47 & 85.23 & 86.38 & 22.88 &  6.28 & 17.34 \\
\bottomrule
\end{tabular}
\vspace{0.5ex}
\begin{flushleft}
\scriptsize{\textit{Note:} Values shown as percentages ({\%} symbol omitted). T0–T3 = Tasks 0–3; 
Avg = Average Performance; W.Risk = Worst Risk; SD = Standard Deviation; Rng = Range.}
\end{flushleft}
\end{table}

\begin{figure}[ht]
    \centering
    \includegraphics[
      width=0.9\textwidth,          % still allow it to span the page width…
      height=0.5\textheight,     % …but never exceed half the text height
      keepaspectratio            % preserve the original aspect ratio
    ]{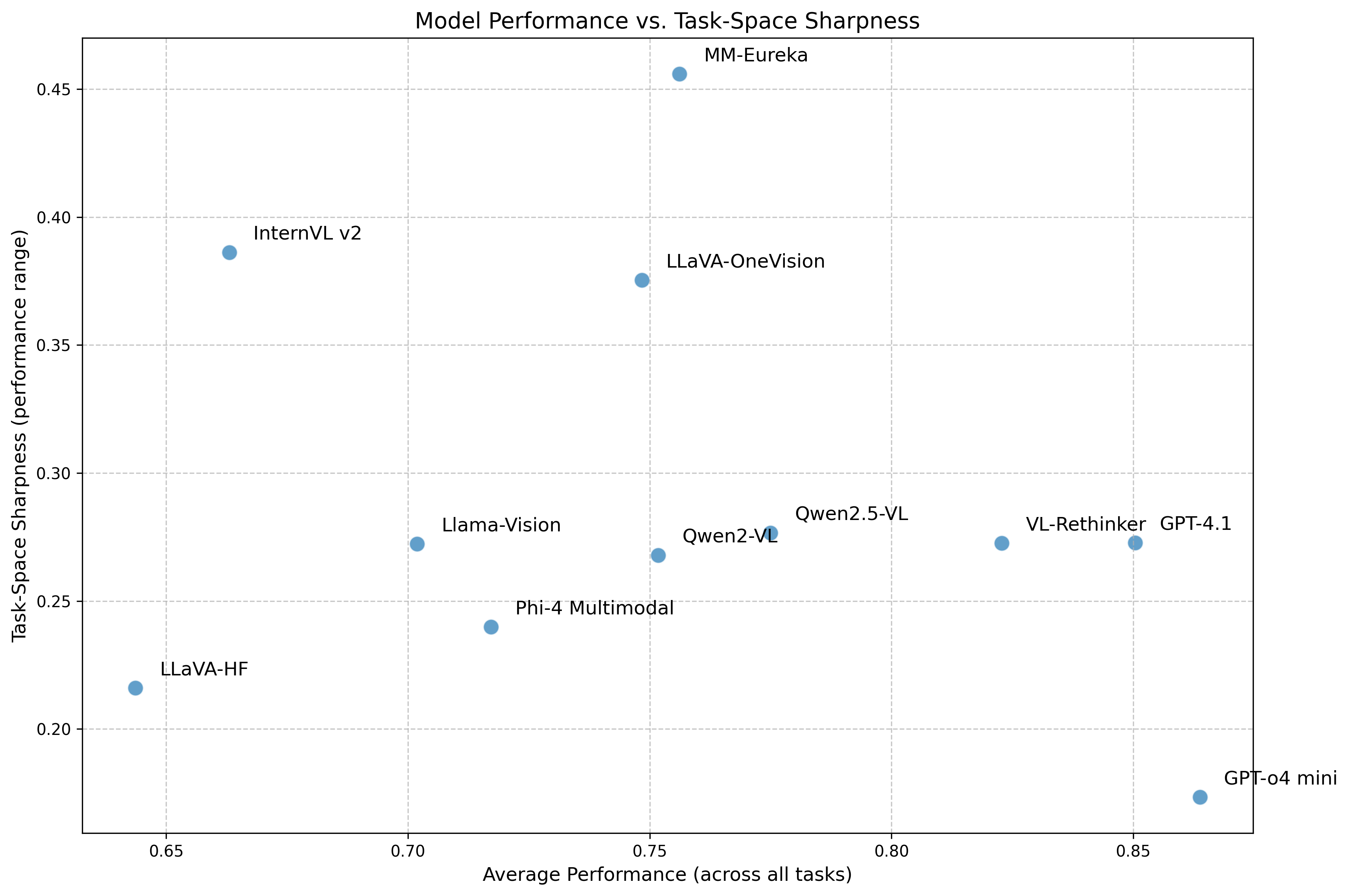}
    \caption{How performance align with sharpness for RealworldQA dataset?}
    \label{fig:sharpness_vs_performance_realworldqa}
\end{figure}

\begin{figure}[ht]
    \centering
    \includegraphics[width=0.9\textwidth]{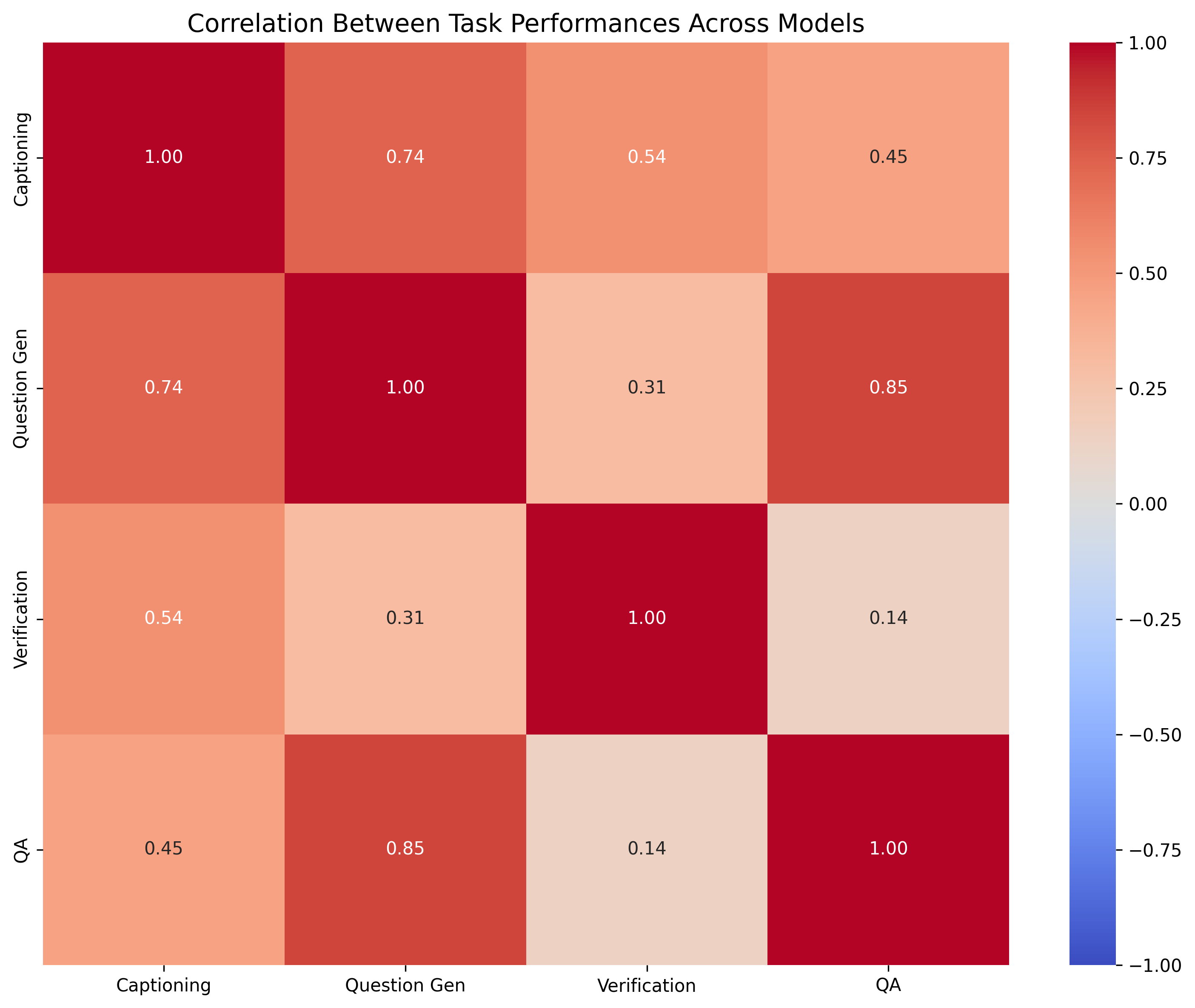}
    \caption{How related are each task for realworldQA dataset?}
    \label{fig:task_correlation_realworldqa}
\end{figure}

\newpage
\subsection{Results for CVRR-ES dataset}

\begin{table}[h]
\centering
\caption{Performance of MLLMs Across Four Tasks on CVRR-ES dataset.}
\label{tab:cvrr_es}
\renewcommand{\arraystretch}{1.0}
\begin{tabular}{l r r r r | r r r r}
\toprule
\textbf{Model} & \textbf{T0} & \textbf{T1} & \textbf{T2} & \textbf{T3} & \textbf{Avg} & \textbf{W.Risk} & \textbf{SD} & \textbf{Rng} \\
\midrule
GPT-4.1            & 83.06 & 80.24 & 76.65 & 85.81 & 81.44 & 23.35 &  3.39 &  9.16 \\
GPT-o4 mini        & 83.89 & 61.32 & 77.52 & 81.39 & 76.03 & 38.68 &  8.79 & 22.57 \\
InternVideo2.5     & 72.00 & 69.30 & 82.58 & 61.07 & 71.24 & 38.93 &  7.69 & 21.51 \\
InternVL v2        & 71.81 & 72.53 & 64.67 & 50.40 & 64.85 & 49.60 &  8.89 & 22.13 \\
LLaVA-OneVision    & 81.76 & 72.03 & 67.27 & 58.91 & 69.99 & 41.09 &  8.26 & 22.86 \\
Phi-4 MM           & 70.50 & 62.19 & 58.95 & 55.78 & 61.86 & 44.22 &  5.48 & 14.72 \\
Qwen2.5-VL         & 80.12 & 65.93 & 65.77 & 59.57 & 67.85 & 40.43 &  7.54 & 20.55 \\
Qwen2.5-VL (7B)    & 84.44 & 76.19 & 74.03 & 78.03 & 78.17 & 25.97 &  3.89 & 10.41 \\
Qwen2-VL           & 78.86 & 64.01 & 65.93 & 45.35 & 63.54 & 54.65 & 11.95 & 33.51 \\
\midrule[\lightrulewidth]
VideoChat-R1       & 82.23 & 73.44 & 87.46 & 74.48 & 79.41 & 26.56 &  5.76 & 14.02 \\
Video-R1           & 82.35 & 63.52 & 74.42 & 67.55 & 71.96 & 36.48 &  7.15 & 18.83 \\
\bottomrule
\end{tabular}
\vspace{0.5ex}
\begin{flushleft}
\scriptsize{\textit{Note:} Values shown as percentages ({\%} symbol omitted). T0–T3 = Tasks 0–3; 
Avg = Average Performance; W.Risk = Worst Risk; SD = Standard Deviation; Rng = Range.}
\end{flushleft}
\end{table}

\begin{figure}[ht]
  \centering
  \includegraphics[
    width=0.9\textwidth,
    height=0.5\textheight,
    keepaspectratio
  ]{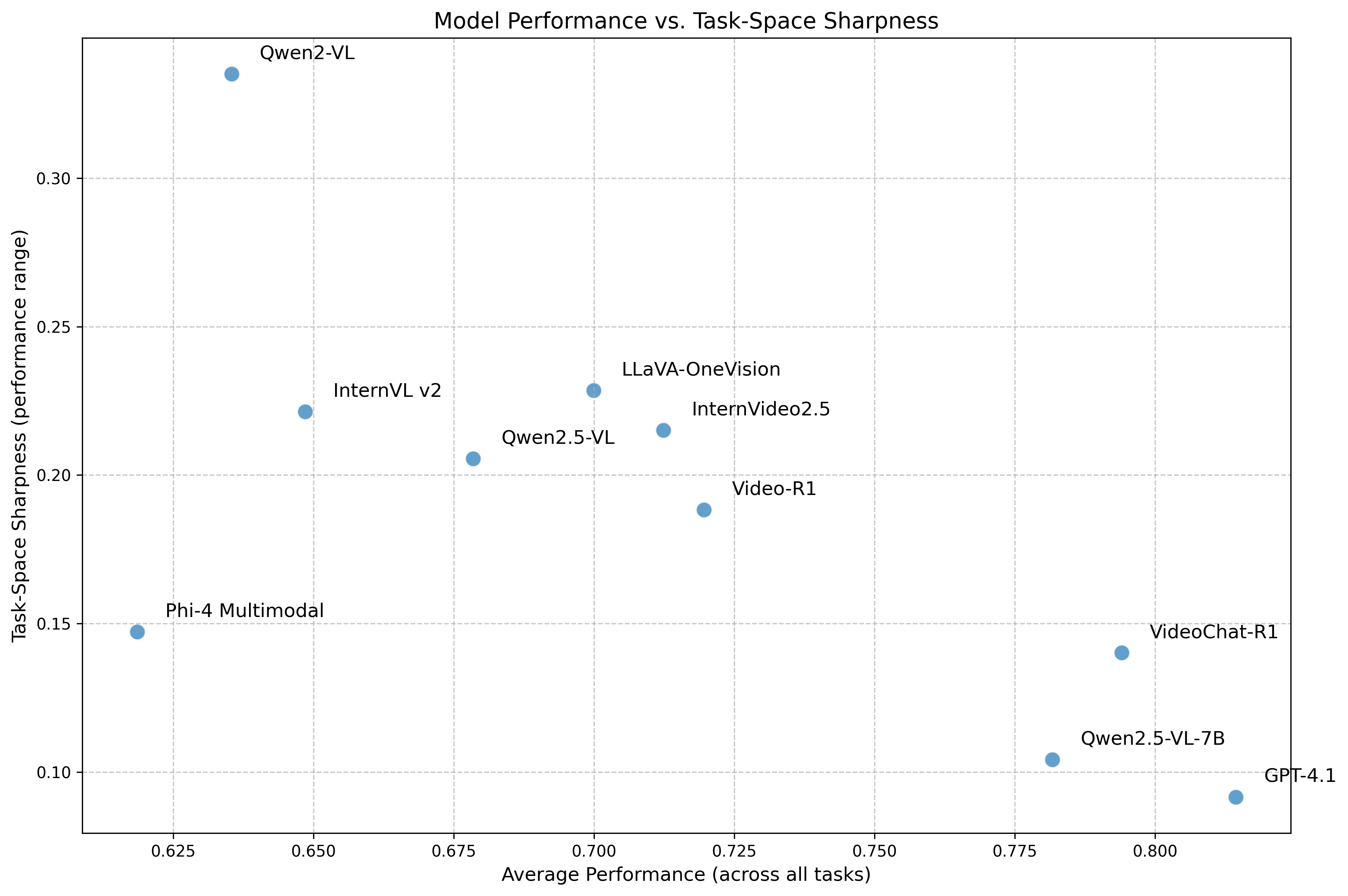}
  \caption{How does performance align with sharpness on the CVRR-ES dataset?}
  \label{fig:sharpness_vs_performance_cvrr_es}
\end{figure}

%%%%%%%%%%%%%%%%%%%%%%%%%%%%%%%%%%%%%%%%%%%%%%%%%%%%%%%%%%%%
\clearpage

\newpage

\end{document}